\documentclass[pdflatex,sn-mathphys-num]{sn-jnl}

\usepackage{graphicx}%
\usepackage{multirow}%
\usepackage{amsmath,amssymb,amsfonts}%
\usepackage{amsthm}%
\usepackage{mathrsfs}%
\usepackage[title]{appendix}%
\usepackage{xcolor}%
\usepackage{textcomp}%
\usepackage{manyfoot}%
\usepackage{booktabs}%
\usepackage{algorithm}%
\usepackage{algorithmicx}%
\usepackage{algpseudocode}%
\usepackage{listings}%

\theoremstyle{thmstyleone}%
\theoremstyle{thmstyletwo}%

\theoremstyle{thmstylethree}%

\begin{document}

\title[Article Title]{DiffFit: Disentangled Garment Warping and \\
Texture Refinement for Virtual Try-On}


\author*[1]{\fnm{Xiang} \sur{Xu}}\email{xu@cjlu.edu.cn}



\affil*[1]{\orgname{China Jiliang  University}}




\abstract{Virtual try-on (VTON) aims to synthesize realistic images of a person wearing a target garment, with broad applications in e-commerce and digital fashion. While recent advances in latent diffusion models have substantially improved visual quality, existing approaches still struggle with preserving fine-grained garment details, achieving precise garment-body alignment, maintaining inference efficiency, and generalizing to diverse poses and clothing styles.
To address these challenges, we propose DiffFit, a novel two-stage latent diffusion framework for high-fidelity virtual try-on. DiffFit adopts a progressive generation strategy: the first stage performs geometry-aware garment warping, aligning the garment with the target body through fine-grained deformation and pose adaptation. The second stage refines texture fidelity via a cross-modal conditional diffusion model that integrates the warped garment, the original garment appearance, and the target person image for high-quality rendering.
By decoupling geometric alignment and appearance refinement, DiffFit effectively reduces task complexity and enhances both generation stability and visual realism. It excels in preserving garment-specific attributes such as textures, wrinkles, and lighting, while ensuring accurate alignment with the human body.
Extensive experiments on large-scale VTON benchmarks demonstrate that DiffFit achieves superior performance over existing state-of-the-art methods in both quantitative metrics and perceptual evaluations.}




\maketitle

\section{Introduction}\label{sec1}

As the e-commerce industry continues to expand rapidly, enhancing user experience through interactive and personalized content has become crucial for improving customer engagement and conversion rates. In this context, virtual try-on (VTON)\cite{wang2018toward,han2018viton} has emerged as a transformative technology, allowing users to visualize themselves wearing selected garments directly from product images. By transferring a clothing item onto a target person image, VTON enables personalized virtual fitting, thereby bridging the gap between physical and digital shopping. 

While VTON and virtual dressing (VD) \cite{shen2025imagdressing,chen2024MagicClothing} both focus on clothing-body integration, their goals and application scenarios differ. VTON primarily targets end-user personalization, supporting garment fit visualization on specific individuals. In contrast, VD is designed for merchant-facing scenarios, enabling large-scale clothing display across diverse body types, poses, and styles. VTON emphasizes static but personalized try-on experiences, whereas VD focuses on scalable, dynamic garment generation across varied conditions. Recent works in layout- and attribute-controllable generation \cite{shen2025imagharmony} have further highlighted the potential of structured conditioning in improving visual fidelity and controllability in garment synthesis.

Despite recent progress, achieving photorealistic and semantically aligned try-on results remains challenging. Precise garment deformation and fitting require accurate modeling of pose-dependent geometry, preservation of fine-grained garment details such as textures and wrinkles, and robustness to occlusions and diverse body shapes. Existing methods often struggle with realistic fabric draping and local detail preservation under complex pose conditions. The problem is further exacerbated by the difficulty of maintaining semantic coherence between the original garment and the synthesized try-on image.
Early VTON approaches \cite{bai2022single,ge2021parser,lee2022high,morelli2022semantic} primarily relied on generative adversarial networks (GANs), employing warping modules to align garment features and generators to synthesize final images. While effective to some extent, GAN-based methods suffer from optimization instability and limited capacity to preserve high-frequency textures. This leads to blurry outputs and poor garment-body alignment, especially in the presence of occlusions or pose variations. These limitations have motivated the shift toward more stable and expressive generative paradigms.

Recently, latent diffusion models (LDMs) \cite{rombacher2022high}
have demonstrated remarkable capabilities in controlled image synthesis tasks, including pose-guided person generation \cite{shen2024imagpose} and long-term motion-consistent generation \cite{shen2025long}. Their iterative denoising process allows for better integration of semantic priors and more effective texture preservation. In the VTON domain, diffusion-based methods enable multi-step refinement of garment fitting and detail rendering, overcoming many limitations of adversarial training.

In this work, we propose DiffFit, a two-stage latent diffusion framework for high-fidelity virtual try-on. Our method adopts a progressive generation strategy that decouples geometric alignment and texture refinement into two stages. In the first stage, we introduce a pose-adaptive diffusion network that implicitly learns semantic correspondences between garments and body shapes in the latent space, generating pose-aligned warped garments without relying on explicit keypoint annotations. This design is inspired by prior work on geometry-aware garment modeling \cite{shen2025imaggarment}. In the second stage, we present a cross-modal conditional diffusion network, which integrates the warped garment with VAE-encoded representations of the original garment and the target person. This enables faithful texture reconstruction, occlusion handling, and photorealistic rendering.
We evaluate DiffFit on three widely-used VTON benchmarks: VITON-HD \cite{choi2021vitonhd}, Dress Code \cite{morelli2022semantic}, and IGPair \cite{shen2025imagdressing}. Extensive experiments demonstrate that our approach achieves state-of-the-art performance in both quantitative metrics and perceptual quality. Moreover, cross-domain evaluation shows strong generalization to unseen poses and garment types. Visual comparisons highlight DiffFit’s ability to preserve texture details, generate realistic fabric deformations, and maintain coherent garment-body alignment.
Our main contributions are summarized as follows:
\begin{itemize}
\item We propose {DiffFit}, a novel two-stage latent diffusion framework that progressively models geometric alignment and texture refinement for high-fidelity virtual try-on.
\item We design a {pose-adaptive diffusion network} that establishes garment-body correspondence in the latent space, enabling precise and natural garment fitting across diverse poses.
\item We introduce a {cross-modal fusion module} for conditional rendering that preserves texture consistency, handles occlusions, and synthesizes photorealistic results with fine structural details.
\end{itemize}

\section{Related Work}\label{sec:rw}

\subsection{Virtual Try-on}

Virtual try-on (VTON) has gained significant attention in computer vision due to its wide applicability in digital fashion and personalized e-commerce. Early approaches predominantly leveraged generative adversarial networks (GANs)~\cite{creswell2018generative} as their generative backbone. These methods typically adopt a two-stage pipeline~\cite{bai2022single,ge2021parser,xie2023gpvton,yang2020towards}: a geometric warping stage that deforms garments to align with the target human body, followed by a synthesis stage that fuses the warped garment with the person image.
VITON-HD~\cite{choi2021vitonhd} proposes a synchronized optimization strategy to jointly refine garment deformation and human segmentation, addressing spatial misalignment and occlusion issues. GPVTON~\cite{xie2023gpvton} introduces a dual-path network to separately handle local garment regions and global body structure. While effective, traditional GAN-based methods often struggle with preserving high-frequency garment details, especially in complex poses or backgrounds, and are limited in their ability to generalize to unseen poses or body shapes.

With the emergence of diffusion models, several recent VTON methods have transitioned to more stable and expressive generative paradigms. DCI-VTON~\cite{gou2023taming} integrates diffusion modules with conventional garment warping to improve texture preservation. LaDI-VTON~\cite{morelli2023ladivton} introduces enhanced skip connections to mitigate information loss during autoencoding. StableVITON~\cite{kim2024stableviton} removes the need for explicit warping by learning garment-to-body correspondence directly in the latent space of a pre-trained diffusion model. However, these approaches still face challenges related to semantic alignment, fine-grained detail preservation, and inference efficiency.
To address these limitations, multi-branch and modular frameworks have been explored. TryOnDiffusion~\cite{zhu2023TryOnDiffusion} employs dual UNets to separately model garment and person features. OOTDiffusion~\cite{xu2024bOOTDiffusion} incorporates attention-based fusion to enhance semantic coherence. IMAGDressing-v1~\cite{shen2025imagdressing} introduces a garment-specific UNet alongside a denoising UNet with hybrid attention, effectively balancing garment appearance and conditional controls. Our work builds upon these insights by decoupling pose-guided garment warping and texture-aware rendering within a unified latent diffusion framework, achieving better alignment and visual fidelity across diverse VTON scenarios.

\subsection{Latent Diffusion Models}

Latent diffusion models (LDMs)~\cite{rombacher2022high} have revolutionized image generation by combining variational autoencoding with multi-step denoising in a compressed latent space. This design drastically reduces computational cost while retaining rich semantic representations. Specifically, an input image $x$ is encoded into a latent variable $z$ via a variational autoencoder (VAE) encoder $\mathcal{E}$, and the reverse process reconstructs $\hat{x} = \mathcal{D}(z)$. The diffusion process gradually corrupts $z$ by injecting Gaussian noise over $T$ steps:
\begin{equation}
    z_t = \sqrt{\bar{\alpha}_t} z_0 + \sqrt{1 - \bar{\alpha}_t} \epsilon, \qquad \epsilon \sim \mathcal{N}(0, I),
\end{equation}
and the model is trained to predict the noise:
\begin{equation}
L_{LDM} = \mathbb{E}_{\mathbf{z}_t, \epsilon, t}\left\|\epsilon_\theta\left(\mathbf{z}_t, t\right)-\epsilon\right\|^2.
\end{equation}
LDMs have shown remarkable generative quality and controllability in diverse vision tasks. In VTON, they enable flexible feature composition and high-fidelity rendering. For instance, TryOnDiffusion~\cite{zhu2023TryOnDiffusion} and StableVITON~\cite{kim2024stableviton} demonstrate that pose and garment can be integrated in the latent space without explicit spatial warping.

Beyond VTON, recent works have extended LDMs to more complex generative tasks, such as pose-guided human synthesis~\cite{shen2023advancing}, story-consistent image generation~\cite{shen2025boosting}, and stylized character generation~\cite{gao2025faceshot}. These methods explore progressive conditioning, cross-modal fusion, and motion-guided priors to achieve enhanced coherence and structure preservation. Notably, the IMAGDressing~\cite{shen2025imagdressing} framework exemplifies how hybrid attention and garment-specific modules can enhance controllability in fashion generation. Similarly, warping-guided latent diffusion approaches~\cite{gao2024exploring} demonstrate that integrating explicit geometric priors into the diffusion process improves fitting accuracy and detail alignment.
Moreover, methods like Ensembling Diffusion~\cite{wang2024ensembling} further show that adaptive feature aggregation across multiple diffusion models can boost consistency and robustness—a principle that aligns with our progressive architecture design. Inspired by these advances, our proposed \textbf{DiffFit} model leverages pose-aware latent guidance and cross-modal fusion to generate photorealistic and structurally aligned try-on results across a wide range of poses and garments.

\section{Proposed Method}\label{sec:method}

\subsection{Overview}

\begin{figure*}[t]
  \centering
  \includegraphics[width=1\textwidth]{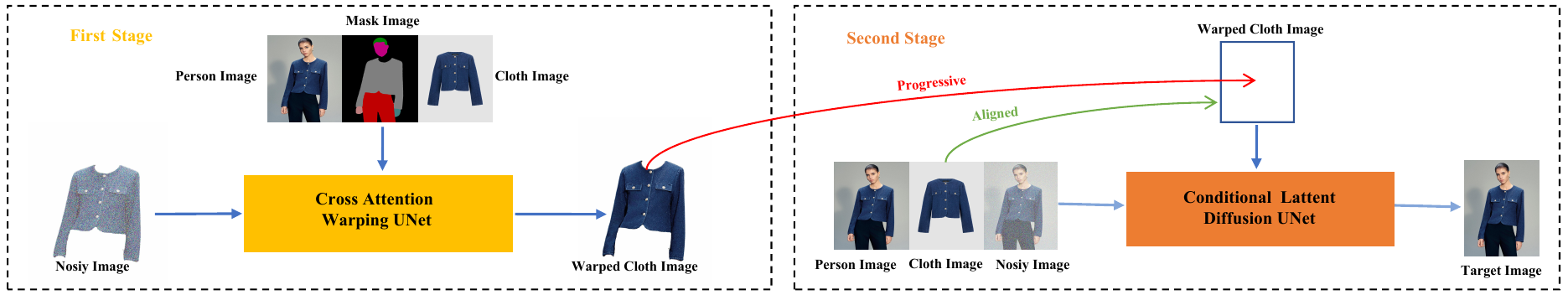}
  \caption{Overview of the proposed two-stage DiffFit framework. Stage I warps the garment to fit the target body using cross-attention-based diffusion, and Stage II fuses the warped garment with the person image via a conditional latent diffusion model to generate the final  high-fidelity try-on result.}
  \label{fig:0pdf}
\end{figure*}

To address the challenges of semantic misalignment and detail preservation in virtual try-on, we propose a novel two-stage latent diffusion framework.  An overview of our proposed model is illustrated in Figure~\ref{fig:0pdf}. Our pipeline is designed to explicitly disentangle garment-body correspondence from high-fidelity image synthesis, thereby improving both geometric fitting and texture realism.

Given a person image $x_p \in \mathbb{R}^{H \times W \times 3}$, a clothing image $x_c \in \mathbb{R}^{H \times W \times 3}$, and a binary mask $x_m \in \mathbb{R}^{H \times W \times 1}$ indicating the body region to be replaced, our goal is to generate a photorealistic try-on image $\hat{x} \in \mathbb{R}^{H \times W \times 3}$ in which the garment is realistically worn by the target person.

Our framework consists of the following two stages:
\begin{itemize}
    \item \textbf{Stage I: Garment Warping.} We employ a cross-attention-based diffusion model to generate a warped garment image $\tilde{x}_c$ that is semantically aligned with the target pose and body shape, while preserving garment texture and structure.
    \item \textbf{Stage II: Try-on Synthesis.} The warped garment $\tilde{x}_c$, original person image $x_p$, and reference garment $x_c$ are fused via a conditional latent diffusion model to synthesize the final try-on result $\hat{x}$, ensuring seamless integration and high-fidelity detail.
\end{itemize}

By directly utilizing image inputs for both garment and target person, our method circumvents the limitations of text-guided editing, which often fails to capture subtle appearance cues. The two-stage design also allows for explicit control over garment deformation and appearance consistency, as motivated by our observations in Section \ref{sec:rw}.

\subsection{Cross Attention Warping UNet (CAW-UNet)}

\begin{figure*}[t]
  \vspace{-10pt}
  \begin{minipage}[t]{0.49\textwidth}
    \centering
    \includegraphics[width=\linewidth]{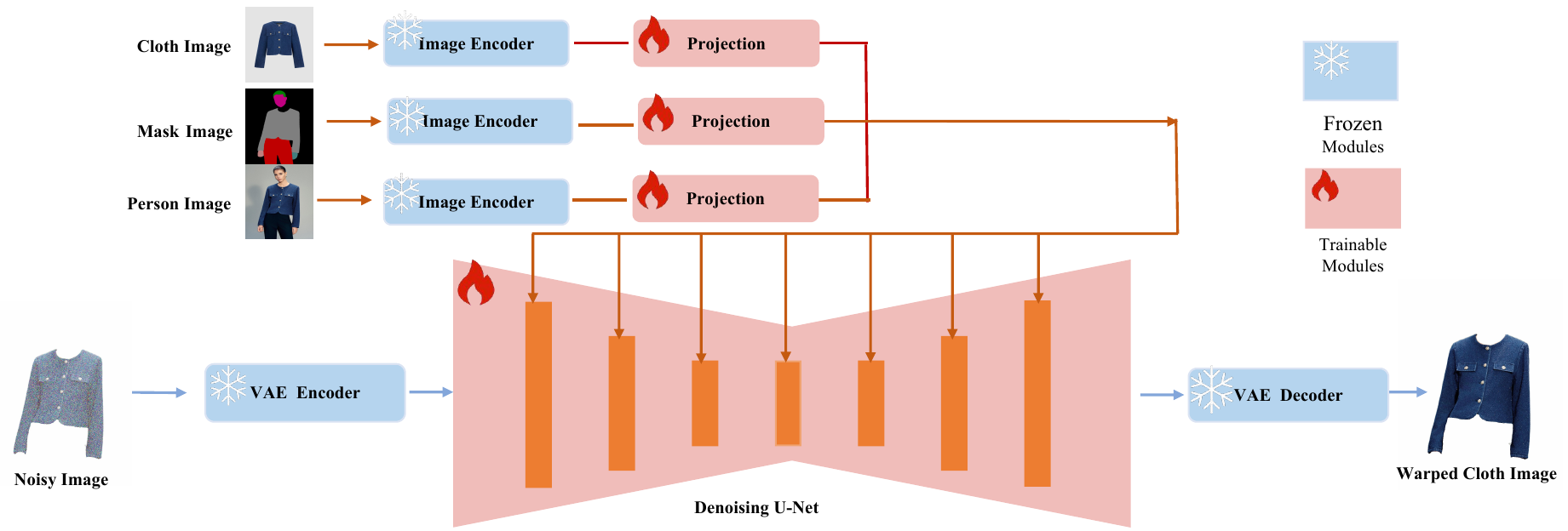}
    \vspace{-10pt}
    \caption{Cross attention warping UNet (CAW-UNet) architecture. Tokens of person, cloth, and mask images are extracted via a frozen image encoder and a trainable projection layer, and subsequently injected into the UNet through cross-attention mechanism.}
    \label{fig:caw_unet}
  \end{minipage}
  \hfill
  \begin{minipage}[t]{0.49\textwidth}
    \centering
    \includegraphics[width=\linewidth]{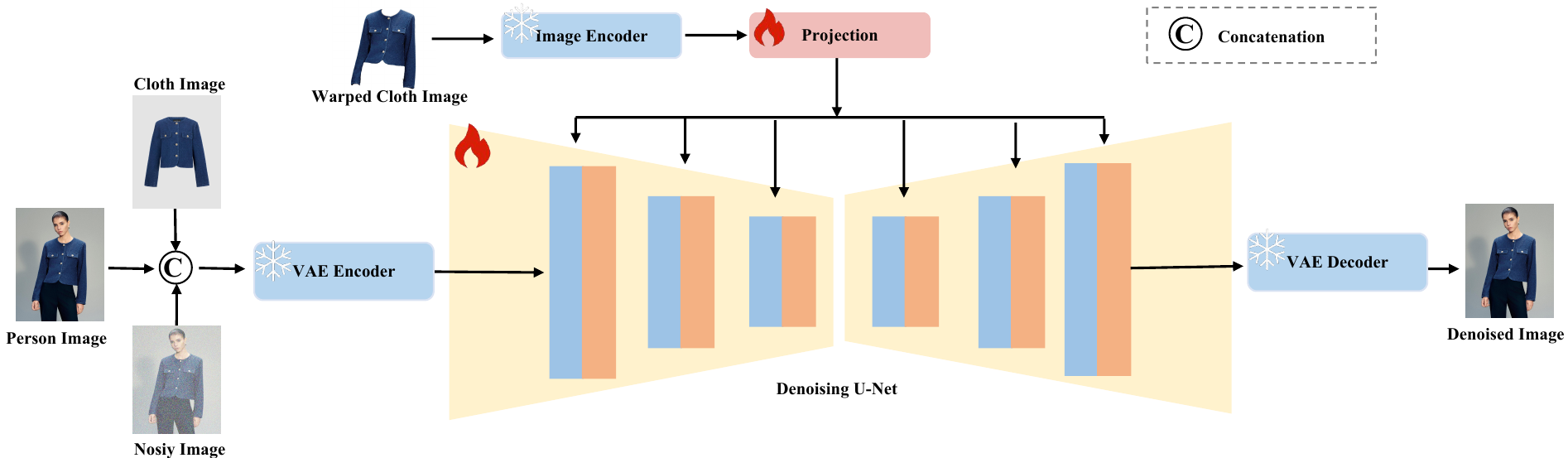}
    \vspace{-10pt}
    \caption{Overview of the second stage of our virtual try-on framework. The person image, the original garment reference image and the warped garment image generated from the first stage are jointly fed into a trainable U-Net. By effectively aggregating multi-source features, the network synthesizes the final try-on result. }
    \label{fig:two-stage}
  \end{minipage}
  \vspace{-10pt}
\end{figure*}

A key challenge for virtual try-on is to accurately align the garment with the target body while preserving the garment's fine-grained features. To this end, we introduce the cross attention warping UNet (CAW-UNet), which leverages multi-source feature extraction and cross-attention fusion, as shown in Figure \ref{fig:caw_unet}.

Given $x_p$, $x_c$, and $x_m$, we extract their visual features using fixed image encoders $E_p$, $E_c$, and $E_m$, producing feature maps $f_p = E_p(x_p)$, $f_c = E_c(x_c)$, and $f_m = E_m(x_m)$. These feature maps are subsequently projected into a unified embedding space via a trainable Q-Former~\cite{li2023blip}, resulting in token sequences $t_p = Q(f_p)$, $t_c = Q(f_c)$, and $t_m = Q(f_m)$, where $Q(\cdot)$ denotes the Q-Former projection.

The concatenated token sequence $[{t}_p; {t}_c; {t}_m]$ is injected into the denoising UNet via cross-attention layers, enabling the model to dynamically attend to garment, person, and mask features during the diffusion process. This design allows for flexible and accurate warping of the garment according to the target body pose, while maintaining texture fidelity.

The output of the first stage is a warped garment image $\tilde{x}_c$, which serves as an intermediate representation for the subsequent synthesis stage.

\subsection{Conditional Input and Cross-Modal Fusion in UNet}

In the second stage, our objective is to seamlessly integrate the warped garment with the person image, while preserving both garment details and human identity. To achieve this, we design a conditional latent diffusion UNet with cross-modal fusion, as shown in Figure \ref{fig:two-stage}.

Let $z_p = \mathcal{E}(x_p)$, $z_c = \mathcal{E}(x_c)$, and $z_{\tilde{c}} = \mathcal{E}(\tilde{x}_c)$ denote the latent representations of the person image, original garment, and warped garment, respectively, where $\mathcal{E}$ is a frozen VAE encoder. These latents are concatenated along the channel dimension and fed into the U-Net backbone. To accommodate the increased input dimensionality, the first convolutional layer is modified accordingly.

Additionally, the feature embedding of the warped garment, extracted via a fixed image encoder and a trainable projection layer, is injected into the intermediate layers of the U-Net through cross-attention. This mechanism enables the model to adaptively fuse garment appearance, warping information, and human context, thereby enhancing the realism and consistency of the generated try-on image.

The inclusion of both original and warped garment features provides strong constraints for preserving garment patterns and textures, while the person image ensures the retention of identity and pose.

\subsection{Loss Function}

To optimize the proposed framework, we adopt a two-stage training objective that follows the standard diffusion model paradigm, conditioning on multi-modal inputs in a unified manner.

Let $t_p$, $t_c$, and $t_m$ denote the learned conditioning token embeddings for the person, clothing, and mask images, respectively. In the first stage (garment warping), the model is trained to align the garment with the target body, using the following objective:

\begin{equation}
L_{1} = \mathbb{E}_{\mathbf{z}_t, \epsilon, t, t_p, t_c, t_m} \left\| \epsilon_\theta(\mathbf{z}_t, t, t_p, t_c, t_m) - \epsilon \right\|^2,
\end{equation}
where $\mathbf{z}_t$ denotes the noisy latent variable at diffusion timestep $t$, $\epsilon$ is Gaussian noise sampled from $\mathcal{N}(\mathbf{0}, \mathbf{I})$, and $\epsilon_\theta$ is the denoising network parameterized by $\theta$.

In the second (synthesis) stage, the goal is to seamlessly integrate the warped garment with the person image for realistic try-on results. Let $t_{wc}$ denote the embedding of the warped garment. The loss function for this stage is defined as:
\begin{equation}
L_{2} = \mathbb{E}_{\mathbf{z}_t, \epsilon, t, t_p, t_c, t_{wc}} \left\| \epsilon_\theta(\mathbf{z}_t, t, t_p, t_c, t_{wc}) - \epsilon \right\|^2,
\end{equation}
where all variables retain the same meanings as above.

This two-stage training strategy ensures that the model can learn both precise garment-body alignment and high-fidelity image synthesis, ultimately producing photorealistic and semantically aligned virtual try-on results.

\section{Experiment and Analysis}\label{sec:exp}

We evaluate the effectiveness of our proposed method by conducting comprehensive comparisons with several state-of-the-art virtual try-on approaches, including LaDI-VTON~\cite{morelli2023ladivton}, DCI-VTON~\cite{gou2023taming}, and \mbox{IMAGEDressing}~\cite{shen2025imagdressing}. For all comparative methods, we use official pre-trained weights when available; otherwise, we retrain the models using the released code and recommended settings.

\subsection{Datasets}

We conduct experiments on three widely used virtual try-on datasets: DressCode~\cite{morelli2022dress}, VITON-HD~\cite{choi2021vitonhd}, and IGPair~\cite{shen2025imagdressing}.

The DressCode dataset~\cite{morelli2022dress} contains a total of 53,795 high-quality image pairs, covering three apparel categories: upper-body clothing (15,366 pairs), lower-body clothing (8,951 pairs), and dresses (29,478 pairs). All images are standardized to a resolution of $1024 \times 768$ pixels. The dataset is split into 48,395 pairs for training and 5,400 pairs for testing, with the test set uniformly sampled across all categories to support fine-grained performance evaluation.

The VITON-HD dataset~\cite{choi2021vitonhd} is tailored to address challenges of complex human poses and garment alignment in virtual try-on scenarios. It comprises 13,679 image pairs, each consisting of a frontal-view female portrait ($1024 \times 768$ pixels) and the corresponding upper-body garment image. The dataset is divided into 11,647 training pairs and 2,032 testing pairs.

The IGPair dataset~\cite{shen2025imagdressing} includes 324,857 high-quality image pairs, each formed by a garment item and one or more corresponding images of human models wearing that item. Compared with existing benchmarks, IGPair is significantly larger in scale and features images collected at ultra-high resolution, enabling precise preservation and modeling of subtle garment characteristics.

\subsection{Evaluation Metrics}

To ensure a fair comparison with baseline methods, all experiments are conducted at a fixed resolution. For quantitative evaluation, we adopt the Structural Similarity Index ($\mathrm{SSIM}$)~\cite{wang2004image} and Learned Perceptual Image Patch Similarity ($\mathrm{LPIPS}$)~\cite{zhang2018unreasonable} in the paired setting to measure reconstruction quality. To assess the realism and distributional consistency of generated images, the Fréchet Inception Distance ($\mathrm{FID}$)~\cite{heusel2017gans} and Kernel Inception Distance ($\mathrm{KID}$)~\cite{binkowski2018demystifying} are computed in both paired ($\mathrm{FID}_p$, $\mathrm{KID}_p$) and unpaired ($\mathrm{FID}_u$, $\mathrm{KID}_u$) settings.

\subsection{Implementation Details}

All experiments were conducted using a single NVIDIA RTX 4090 GPU. Both stages of our model share the same optimization and training strategy. Specifically, we employ the AdamW optimizer with a fixed learning rate of $5 \times 10^{-5}$. The model is trained for 50,000 steps with a batch size of 6. Input images are uniformly resized to $640 \times 512$ pixels.

For inference, image generation is performed using the UniPC sampler, with the total number of sampling steps set to 50 and a guidance scale parameter $w$ of 7.5. All other hyperparameters are kept at their default values unless otherwise specified.

\subsection{Comparison with State-of-the-Art Methods}

To thoroughly evaluate the effectiveness of our approach, we compare it with several state-of-the-art virtual try-on methods, including GP-VTON~\cite{xie2023gpvton}, VITON-HD~\cite{choi2021vitonhd}, DCI-VTON~\cite{gou2023taming}, and LaDI-VTON~\cite{morelli2023ladivton}.
Quantitative results for IMAGEDressing are not included, as its relevant evaluation metrics have not been publicly released.

Table~\ref{tab:dc} summarizes the quantitative results on the DressCode dataset. Our method achieves the best $\mathrm{FID}_p$ (4.06), $\mathrm{KID}_p$ (1.51), $\mathrm{FID}_u$ (6.28), and the highest $\mathrm{SSIM}$ (0.986), indicating superior image fidelity and structural similarity. For $\mathrm{KID}_u$, our score (1.71) is nearly identical to that of DCI-VTON (1.70), reflecting competitive performance in unpaired metric scenarios. Although DCI-VTON attains the lowest $\mathrm{LPIPS}$ (0.0301), our $\mathrm{LPIPS}$ (0.040) remains highly competitive. Meanwhile, LaDI-VTON reports a higher $\mathrm{FID}_p$ (4.85) and lower $\mathrm{SSIM}$ (0.906) compared to our method, indicating less realism and weaker structural retention on DressCode.

Table~\ref{tab:vitonhd} presents the quantitative comparisons on the VITON-HD dataset. Our approach consistently outperforms previous methods, achieving the lowest $\mathrm{LPIPS}$ (0.041), the highest $\mathrm{SSIM}$ (0.916), as well as the best $\mathrm{FID}_p$ (6.62), $\mathrm{KID}_p$ (1.08, tying with LaDI-VTON), and $\mathrm{FID}_u$ (8.21). For $\mathrm{KID}_u$, DCI-VTON obtains the best result (0.49), while our method achieves a closely comparable score (1.36). Notably, LaDI-VTON achieves a $\mathrm{KID}_p$ of 1.08, matching our method, but is outperformed by our approach in $\mathrm{LPIPS}$ and $\mathrm{SSIM}$, underscoring our superior perceptual and structural consistency.

On VITON-HD, our method achieves substantially higher $\mathrm{SSIM}$ and lower $\mathrm{LPIPS}$ than DCI-VTON, with consistently superior or competitive $\mathrm{FID}$ and $\mathrm{KID}$ scores, indicating improved structural fidelity, perceptual similarity, and image realism. LaDI-VTON attains competitive $\mathrm{FID}_p$ and $\mathrm{KID}_p$ values, but is outperformed by our approach in $\mathrm{SSIM}$ and $\mathrm{LPIPS}$. On DressCode, our method also ranks best in $\mathrm{SSIM}$, $\mathrm{FID}_p$, $\mathrm{KID}_p$, and $\mathrm{FID}_u$, and delivers comparable $\mathrm{LPIPS}$ to DCI-VTON, highlighting its robustness across datasets. Overall, our results confirm the effectiveness of our approach in both perceptual quality and structural preservation compared to state-of-the-art baselines.

\begin{table}[t]
\caption{Quantitative comparisons on the DressCode dataset. Best results are highlighted in bold.}
\label{tab:dc}
\centering
\begin{tabular}{lcccccc}
\toprule
Method &
$\mathrm{LPIPS} \downarrow$ & 
$\mathrm{SSIM} \uparrow$ & 
$\mathrm{FID}_p \downarrow$ & 
$\mathrm{KID}_p \downarrow$ & 
$\mathrm{FID}_u \downarrow$ & 
$\mathrm{KID}_u \downarrow$ \\
\midrule
GP-VTON~\cite{xie2023gpvton} & 0.0359 & 0.9479 & -- & -- & 11.98 & -- \\
DCI-VTON~\cite{gou2023taming} & \textbf{0.0301} & 0.948 & -- & -- & 6.48 & \textbf{1.70} \\
LaDI-VTON~\cite{morelli2023ladivton} & 0.064 & 0.906 & 4.85 & 1.61 & 7.50 & 2.83 \\
\midrule
\textbf{Ours} & 0.040 & \textbf{0.986} & \textbf{4.06} & \textbf{1.51} & \textbf{6.28} & 1.71 \\
\bottomrule
\end{tabular}
\footnotetext{Lower values ($\downarrow$) are better for $\mathrm{LPIPS}$, $\mathrm{FID}$, and $\mathrm{KID}$; higher values ($\uparrow$) are better for $\mathrm{SSIM}$. Subscript $p$: paired; $u$: unpaired; --: unavailable.}
\end{table}

\begin{table}[t]
\caption{Quantitative comparisons on the VITON-HD dataset. Best results are highlighted in bold.}
\label{tab:vitonhd}
\centering
\begin{tabular}{lcccccc}
\toprule
Method &
$\mathrm{LPIPS} \downarrow$ & 
$\mathrm{SSIM} \uparrow$ & 
$\mathrm{FID}_p \downarrow$ & 
$\mathrm{KID}_p \downarrow$ & 
$\mathrm{FID}_u \downarrow$ & 
$\mathrm{KID}_u \downarrow$ \\
\midrule
GP-VTON~\cite{xie2023gpvton} & 0.0799 & 0.8939 & -- & -- & 9.197 & -- \\
VITON-HD~\cite{choi2021vitonhd} & 0.116 & 0.863 & 11.01 & 3.71 & 12.96 & 4.09 \\
DCI-VTON~\cite{gou2023taming} & 0.043 & 0.896 & 8.09 & 2.80 & 8.23 & \textbf{0.49} \\
LaDI-VTON~\cite{morelli2023ladivton} & 0.091 & 0.876 & 6.66 & 1.08 & 9.41 & 1.60 \\
\midrule
\textbf{Ours} & \textbf{0.041} & \textbf{0.916} & \textbf{6.62} & \textbf{1.08} & \textbf{8.21} & 1.36 \\
\bottomrule
\end{tabular}
\footnotetext{Lower values ($\downarrow$) are better for $\mathrm{LPIPS}$, $\mathrm{FID}$, and $\mathrm{KID}$; higher values ($\uparrow$) are better for $\mathrm{SSIM}$. Subscript $p$: paired; $u$: unpaired; --: unavailable.}
\end{table}

\begin{figure*}[t]
  \centering
  \includegraphics[width=1\textwidth]{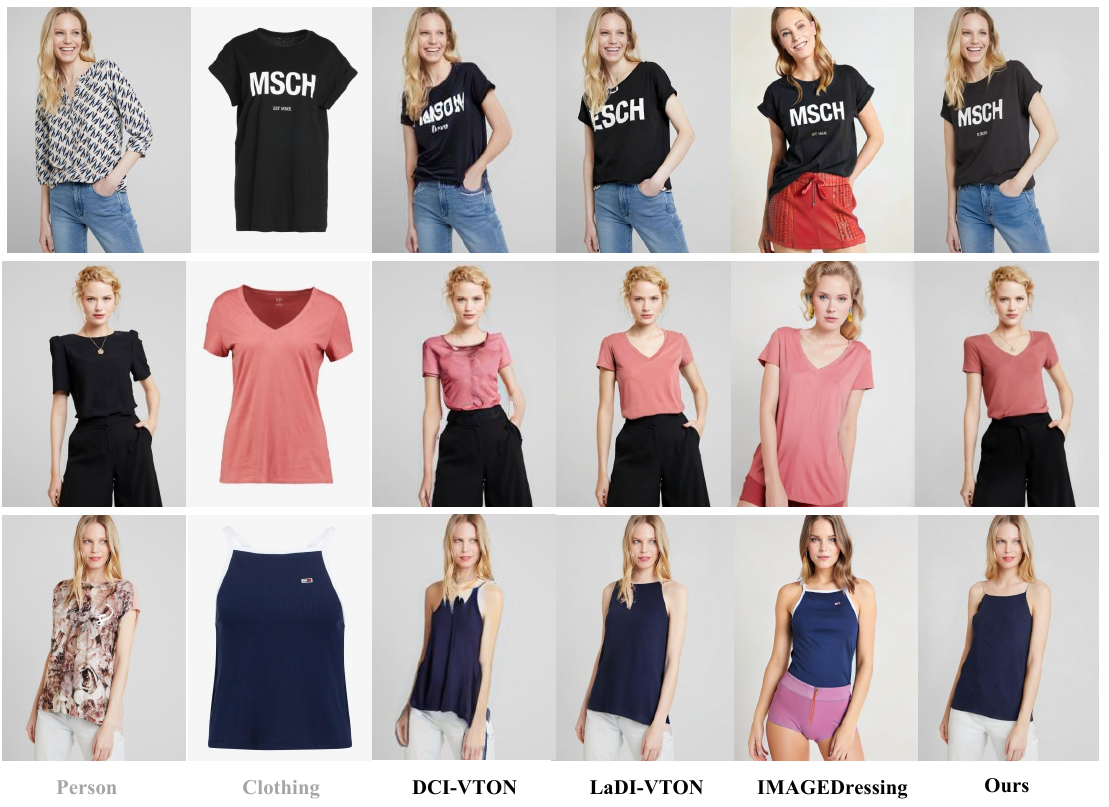}
  \caption{Qualitative comparison of our method with state-of-the-art virtual try-on approaches (DCI-VTON, LaDI-VTON, IMAGEDressing) on challenging cases from the DressCode and VITON-HD datasets.}
  \label{fig:3pdf}
\end{figure*}

For qualitative comparison, we further evaluate the image synthesis quality of our method against three representative approaches, as illustrated in Figure~\ref{fig:3pdf}. Our method demonstrates notable advantages in several key aspects.
First, as observed in the first row, DCI-VTON produces artifacts around the clothing neckline, and when the clothing logo is partially occluded by hair, both LaDI-VTON and DCI-VTON fail to accurately reconstruct the textual details. In contrast, our method successfully restores the logo. Additionally, IMAGEDressing fails to generate hair that matches the model’s appearance.
Second, in the second row, DCI-VTON results in garment deformation near the neckline, while IMAGEDressing is unable to preserve the original body pose. By comparison, our method maintains both the correct pose and garment structure.
Third, as shown in the third row, DCI-VTON fails to retain the intended clothing style, LaDI-VTON produces inconsistent garment positioning, and IMAGEDressing introduces prominent white edges around the vest. In contrast, our results remain visually realistic and structurally coherent across all examples.
These improvements highlight the enhanced capacity of our model to handle complex textures, occlusions, and pose variations in virtual try-on applications.

\subsection{User Study}

To further assess the perceptual effectiveness and practical applicability of our approach, we conducted a comprehensive user study involving comparisons with two state-of-the-art baselines, DCI-VTON and LaDI-VTON, on the DressCode and VITON-HD datasets.

The evaluation was structured around three principal criteria: (1) overall image realism, (2) garment detail preservation, and (3) consistency with human body pose. In each trial, participants were presented with a pair of images: one produced by our method and the other by a baseline. Both the order of images and the choice of baseline were randomized to minimize presentation and expectation biases. Participants were instructed to select the image that appeared more realistic and visually satisfactory according to the specified criterion.

\begin{table}[t]
    \centering
    \caption{User study preference (\%) comparing DCI-VTON, LaDI-VTON, and our method on VITON-HD and DressCode datasets for image realism, garment detail, and body consistency. Higher is better.}
    \label{tab:core_metrics}
    \setlength{\tabcolsep}{0.6em}
    \small
    \begin{tabular}{l l c c c}
        \toprule
        \textbf{Dataset} & \textbf{Criterion} & \textbf{DCI-VTON} & \textbf{LaDI-VTON} & \textbf{Ours} \\
        \midrule
        \multirow{3}{*}{VITON-HD~\cite{choi2021vitonhd}}
            & Image Realism              & 18.4 & 21.2 & 60.4 \\
            & Garment Detail             & 17.1 & 18.6 & 64.3 \\
            & Body Consistency           & 18.3 & 18.6 & 63.1 \\
        \midrule
        \multirow{3}{*}{DressCode~\cite{morelli2022dress}}
            & Image Realism              & 18.6 & 19.3 & 62.1 \\
            & Garment Detail             & 20.1 & 18.8 & 61.1 \\
            & Body Consistency           & 20.1 & 19.2 & 60.7 \\
        \bottomrule
    \end{tabular}
\end{table}

We collected a total of 1,000 valid responses from over 50 volunteers with diverse backgrounds, ensuring a broad and representative sample. To further reduce subjective bias, each image pair was evaluated independently by at least five different participants.

The results shown in Table~\ref{tab:core_metrics} demonstrate a significant preference for our method, selected as superior in more than 60\% of trials.

Our approach consistently outperformed both DCI-VTON and LaDI-VTON across all criteria and datasets. These results indicate that, beyond improvements in objective metrics, our method delivers virtual try-on results that are more convincing and appealing to human observers.

\subsection{Ablation Studies and Analysis}

To thoroughly evaluate the contribution of each component within the DiffFit framework, we conduct a series of ablation studies by systematically removing or modifying key modules. These experiments rigorously assess the impact of architectural design and feature extraction strategies on overall model performance.

\textbf{Feature Extraction Mechanism.}
We investigate the effect of alternative feature extraction strategies in both stages of our framework. Specifically, we compare our proposed projection-based approach with a baseline that utilizes only a fixed image encoder, omitting the projection layer:
\begin{itemize}
    \item Projection-based: This configuration combines a fixed image encoder ($E$) with a learnable projection layer ($Q$) to map clothing image features into a unified embedding space, producing 768-dimensional semantic tokens ($t_c$). These are injected via cross-attention in both the garment warping and the cross-modal fusion UNet modules.
    \item w/o Projection-based: This variant uses only the fixed image encoder ($E$), directly injecting the extracted features into the corresponding UNet modules via cross-attention, without the projection layer.
\end{itemize}

As reported in Table~\ref{tab:ablation_vitonhd_all}, the projection-based method consistently outperforms the baseline across all evaluation metrics. Specifically, our method achieves a lower $\mathrm{LPIPS}$ of 0.041, a higher $\mathrm{SSIM}$ of 0.916, and improved $\mathrm{FID}_p$ and $\mathrm{KID}_p$ scores (6.62 and 1.56, respectively) compared to the w/o projection-based baseline (0.138, 0.651, 13.01, and 4.12, respectively). Similar improvements are observed in the unpaired garment metrics ($\mathrm{FID}_u$/$\mathrm{KID}_u$ of 8.21/1.36 against 15.12/5.21).

\begin{figure}[t]
\centering
\includegraphics[width=0.95\textwidth]{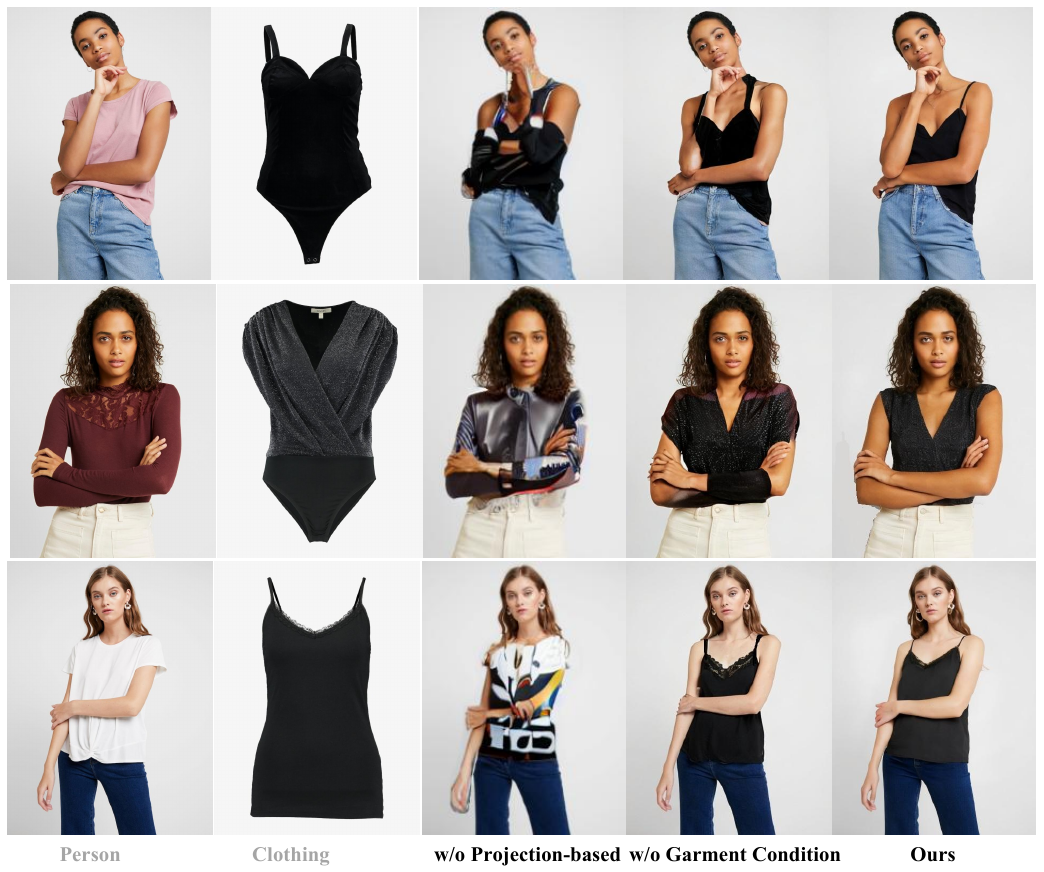}
\caption{Qualitative comparison of ablation settings on virtual try-on. From left to right: input person, reference clothing, w/o projection-based, w/o garment condition, and our full method.}
\label{fig:ablation}
\end{figure}

Figure~\ref{fig:ablation} qualitatively corroborates these findings: in the absence of the projection layer, the generated garments exhibit structural distortions and loss of fine-grained details, while the projection-based approach yields results with superior structural integrity and semantic consistency.

\begin{table}[t]
\centering
\caption{Ablation study on VITON-HD: quantitative results for different settings.}
\label{tab:ablation_vitonhd_all}
\setlength{\tabcolsep}{0.6em}
\begin{tabular}{lcccccc}
\toprule
Ablation Setting
& $\mathrm{LPIPS} \downarrow$
& $\mathrm{SSIM} \uparrow$
& $\mathrm{FID}_p \downarrow$
& $\mathrm{KID}_p \downarrow$
& $\mathrm{FID}_u \downarrow$
& $\mathrm{KID}_u \downarrow$\\
\midrule
w/o Projection-based         & 0.138 & 0.651 & 13.01 & 4.12 & 15.12 & 5.21 \\
w/o Garment Condition        & 0.067 & 0.787 & 8.98 & 2.67 & 9.51 & 1.68 \\
\midrule
Ours                   & 0.041 & 0.916 & 6.62  & 1.56 & 8.21  & 1.36 \\
\bottomrule
\end{tabular}
\end{table}

\textbf{Conditional Input Analysis.}
To further investigate the significance of conditional inputs in the second-stage synthesis module, we conduct an ablation in which the original garment image condition is removed, retaining only the person and warped garment representations (\textit{w/o Garment Condition}). Specifically, the garment image latent ($z_c$) is omitted, whereas the person ($z_p$) and warped garment ($z_{\tilde{c}}$) latents are retained.

Quantitative results presented in Table~\ref{tab:ablation_vitonhd_all} demonstrate that removing the garment condition leads to a clear degradation in performance: SSIM decreases from 0.916 (full model) to 0.787, and FID$_p$ increases from 6.62 to 8.98. LPIPS also increases from 0.041 to 0.067, and KID$_p$ rises from 1.56 to 2.67. Similarly, the unpaired garment FID/KID worsen from 8.21/1.36 to 9.51/1.68. These results collectively indicate that the garment condition is crucial for faithful preservation of garment texture and stylistic consistency in the generated images.

Qualitative comparisons in Figure~\ref{fig:ablation} further corroborate these findings. Without the original garment condition, the synthesized outputs frequently exhibit structural inconsistencies and loss of detail. For instance, in the first example, the garment appears to float above the body, resulting in unrealistic spatial alignment. In the second example, the clothing is unnaturally suspended at the intersection of the arms, leading to ambiguous boundaries between the garment and body parts. In the third example, visual artifacts such as inaccurate lace patterns and misaligned garment straps are observed. By contrast, the full model consistently demonstrates superior structural integrity and semantic coherence, accurately preserving garment localization and intricate details across diverse cases.

\section{Conclusion}\label{sec:con} 

In this paper, we presented DiffFit, a two-stage latent diffusion framework designed to realize photorealistic virtual try-on (VTON). In the first stage, we established an implicit correspondence between the garment and the target body in the diffusion latent space, enabling pose-adaptive garment deformation while preserving structural and texture details. In the second stage, we synthesized the final try-on image by integrating multi-scale features from the warped garment, the original garment appearance, and the person representation using a hybrid UNet architecture.

Experimental results demonstrated that DiffFit advanced the VTON paradigm by combining geometry-aware garment warping with effective multi-conditional fusion, achieving superior geometric alignment and visual realism compared to existing baselines.

Despite these promising results, several limitations remain. For example, the current model is primarily designed for static images and may not generalize well to diverse poses or complex garment types, such as loose or reflective clothing. 

In future work, we plan to explore extensions of DiffFit for dynamic video-based try-on, automatic garment segmentation, and robust performance under real-world variations. Furthermore, we aim to incorporate interactive user control and garment customization, broadening the scope of VTON in practical e-commerce platforms. Through these directions, we hope to further enhance the flexibility, usability, and realism of virtual try-on systems.

\bibliography{main}

\end{document}